\documentclass{article}

 \usepackage[preprint]{neurips_2025}

\usepackage[utf8]{inputenc} 
\usepackage[T1]{fontenc}    
\usepackage{hyperref}       
\usepackage{url}            
\usepackage{booktabs}       
\usepackage{amsfonts}       
\usepackage{nicefrac}       
\usepackage{microtype}      
\usepackage{xcolor}         
\usepackage[most]{tcolorbox}
\usepackage{lmodern}  
\usepackage{inconsolata}  
\usepackage{graphicx}
\usepackage{subcaption}
\usepackage{multirow}

\title{LLMs for Analog Circuit Design Continuum (ACDC)}
\author {
    Yasaman Esfandiari \\HRL Laboratories \And Jocelyn Rego\\HRL Laboratories 
    \And Austin Meyer\\HRL Laboratories \And Jonathan Gallagher\\Stealth \And
    Mia Levy\thanks{mmlevy@hrl.com}\\HRL Laboratories 
}

\begin{document}

\maketitle
\begin{abstract}
  Large Language Models (LLMs) and transformer architectures have shown impressive reasoning and generation capabilities across diverse natural language tasks. 
  However, their reliability and robustness in real-world engineering domains remain largely unexplored, limiting their practical utility in human-centric workflows. 
  In this work, we investigate the applicability and consistency of LLMs for analog circuit design—a task requiring domain-specific reasoning, adherence to physical constraints, and 
  representations—focusing on AI-assisted design where humans remain in the loop. 
  We study how different data representations influence model behavior and compare smaller models (e.g., T5, GPT-2) with larger foundation models (e.g., Mistral-7B, GPT-oss-20B) 
  under varying training conditions. Our results highlight key reliability challenges, including sensitivity to data format, instability in generated designs, 
  and limited generalization to unseen circuit configurations. These findings provide early evidence on the limits and potential of LLMs as tools to enhance human 
  capabilities in complex engineering tasks, offering insights into designing reliable, deployable foundation models for structured, real-world applications.
\end{abstract}


  \section{Problem Statement}
  We are considering analogue circuit layout design problem in which groups of transisters are given to a layout engineer and they need to come up with a layout that not only optimizes for minimal space,
  but also satifies Design Rule Checks (DRC). The goal is to apply LLMs to automate the process. That is, given a netlist which has all the necessary information about the circuit,
  can an LLM generate an optimized layout that passes DRC \& LVS (discussed in Appendix~\ref{app:drc}) .
  In general, a netlist is a plain text representation of the structure of devices and how they interconnect in a given circuit. 
  Each line of the netlist defines a device, and where each pin of that device connects to.
  The layout designer is given the information presented in a netlist, usually in the form of a circuit diagram, 
  and tasked to generate a layout that matches it. 
  This can be a very time consuming task(about $1$ hour per transistor). 
  It can be broken down into 3 stages, each with their own set of challenges discussed in Appendix~\ref{app:circuit_challenges}.
  
  \section{Literature Review}\label{lit}
  Authors in~\cite{gao2021layout} introduce a graph learning framework for automatic symmetry constraint annotation in analog circuits, using an edge-augmented Graph Attention Network (EGAT) 
  to model both node and edge features. While this paper focuses narrowly on learning symmetry annotations via GNNs from circuit netlists, our approach targets a broader rule-aware analog layout 
  generation pipeline—progressing to meet real-world layout constraints and topological structure.
  Authors in~\cite{mirhoseini2021graph} focus on policy learning for macro placement via Reinforcement Learning with transferable GCN representations. Basically, they frame chip floorplanning as a deep reinforcement 
  learning problem, using an edge-based Graph Convolutional Neural Network (GCN) to guide automated placement - achieving layouts in under six hours that match or outperform human-designed ones across 
  key metrics like power, performance, and area.~\cite{zhu2022generative} introduces an ML-guided placement framework that fuses placement and well generation: a GAN produces well regions during analog placement, 
  followed by legalization and refinement, achieving substantial reductions in layout area and half-perimeter wire length (HPWL). While the GAN-based method optimizes physical layout metrics directly through integrated well-aware 
  placement, our framework aspires to broaden representation to enforce DRC and symmetry during layout synthesis across diverse analog design tasks.
  ~\cite{zhu2022automating} review and transition analog layout constraint extraction from traditional heuristic methods toward learning-based approaches, particularly leveraging Graph Neural Networks 
  (e.g., EGAT) that incorporate edge and node features for symmetry and matching constraint detection. While the authors emphasize identifying and extracting layout constraints via GNNs before placing 
  or sizing components, our approach approach focuses on end-to-end generative synthesis of analog layouts, integrating constraint-awareness, spatial reasoning, and real-world applicability in a unified framework. 
  Researchers in~\cite{jeong2025self} propose a UNet-based foundation model trained via self-supervised learning using random patch sampling and masking to bootstrap from limited, unannotated layout data, 
  achieving $96.6\%$ DRC/LVS-clean layout generation across five tasks after fine-tuning. The research conducted in~\cite{krinke2024layout} demonstrate the ability to perform comprehensive DRC and LVS for a commercial 
  analog process (X-FAB XH018 180 nm) using open-source tools—notably KLayout—by automatically generating nearly 74\% of the required design rule scripts from an abstract specification which is a very useful tool for 
  our future work. Additionally, authours~\cite{dhar2020align} published an open-source library which was developed under the DARPA IDEA program that converts unannotated SPICE netlists into GDSII layouts 
  through a structured, hierarchical flow: netlist annotation, parameterized primitive generation (adhering to grid-based design rules), and constraint-driven block placement and routing. It should be noted that 
  ALIGN offers a deterministic, rule-based, hierarchical engine with clear designer entry points and PDK abstractions, whereas our proposed framework aims for a data-driven, multimodal generative pipeline 
  that embeds design rules and structural knowledge within model representations—targeting greater flexibility, potential adaptability to new tasks, and possibly better generalization in real-world analog 
  layout synthesis. Recently, a newer work is published~\cite{mehradfar2025falcon} which offers an end-to-end ML pipeline that transforms performance specifications into analog layouts. Basically, 
  it uses an MLP-driven classifier for topology selection, an edge-centric GNN to predict performance metrics, and gradient-based parameter inference guided by a differentiable layout cost that encodes 
  design rules and parasitic effects, achieving over $99\%$ topology accuracy. Our framework aspires toward a learned generative model that inherently structures analog layout reasoning—combining sequence, 
  graph, and hypergraph representations—for flexible, structure-aware synthesis, potentially better suited for design diversity and real-world layout complexities. The data that FALCON~\cite{mehradfar2025falcon} used 
  is very useful for us in the future when we are fine-tuning multi-modal models.


  \section{Methodology and Results}
  Beyond defining how circuit schematics are represented, it is necessary to formulate layout design as task(s) that a transformer model can digest and solve. 
  We begin by investigating a sub-problem of layout design, framing transistor grouping as a group and subgroup unmasking task. We then train and 
  fine-tune language models to determine the coordinates for transistor placement, starting from a random initial point. Lastly, we finetune a larger 
  transformer to solve the placement task all at once rather than sequentially. A summary of our findings is given in Table~\ref{tab:experiment_summary}.

  \begin{table*}[t]
    \setlength{\tabcolsep}{5pt}
    \centering
    \caption{Summary of experiments conducted across different model groups and configurations.}
    \renewcommand{\arraystretch}{1.05}
    \resizebox{\columnwidth}{!}{
    \begin{tabular}{c c c c c c}
    \hline
    \textbf{Group} & \textbf{Experiment} & \textbf{Data Representation} & \textbf{Model Type} & \textbf{Successful on Synthetic Data} & \textbf{Successful on Real Netlist} \\
    \hline
    \hline
    \multirow{6}{*}{Masking (Training Small Models)} 
     & Toy Problem & Fig~\ref{fig:toy_gpt} & GPT2 & $\times$ & - \\
     & Toy Problem & Fig~\ref{fig:toy_gpt_1D} & GPT2 & $\times$ & - \\
     & Toy Problem & Fig~\ref{fig:toy_gpt_1D} & GPT2+ConvLayer &  \checkmark & - \\
     & Toy Problem & Fig~\ref{fig:toy_gpt_random} & GPT2 &  \checkmark & - \\
     & Subgroup Masking & Fig~\ref{fig:synthetic_netlist} & T5-small & \checkmark & - \\
     & Masking for Layout Generation & Fig~\ref{fig:layout_masking} \& Fig~\ref{fig:grids_t5} & T5-small & \checkmark & -\\
    \hline
    \multirow{2}{*}{Seq-to-Seq (Finetuning LLMS)} 
     & Sequential Layout Generation (data v1) & Fig~\ref{fig:layout_finetune} \&  Fig~\ref{fig:grids_finetune}& Mistral-7B  & \checkmark& - \\
     & Sequential Layout Generation (data v2) & Fig~\ref{fig:promptv2} \&  Fig~\ref{fig:centerpoints} & Mistral-7B  & \checkmark & $\times$ (Fig~\ref{fig:real}) \\
    \hline
    \multirow{1}{*}{All-at-Once (Finetuning LLMS)} 
     & Synthetic Data v3& Fig.~\ref{fig:prompt_v3} \&  Fig~\ref{fig:tr_data_gpt} & GPT-oss-20B & \checkmark & \checkmark (Fig~\ref{fig:real-gpt20})\\
    \hline
    \end{tabular}}
    \caption{Summary of all experiments grouped by training paradigm. 
    Each entry lists the data representation, model type, and whether the approach successfully generated valid layouts on synthetic and/or real netlist data. 
    \checkmark indicates that the method worked effectively for the corresponding data type, while $\times$ indicates failure or inconsistent results}
    \label{tab:experiment_summary}
    \end{table*}

  
  \subsection{Masking}
  
  We look at the layout design problem from a masking lens. The goal here is to see if the transformer is given all the groups in a netlist but one. Can it predict what the missing component should be or 
  where that one component should be placed.
  The hypotesis is that if that's doable, then the transformer can be queried multiple times to generate the whole layout. Formally, we learn the relation of unmasking a circuit layout.
  The ML algorithm learns to fill in holes in a partially filled in design over time given a netlist context.
  At first, we start with filling in a single small, hole, and over time we introduce larger and more holes.
  In the end, we load a netlist as context, and place a series of masks that cover most of the layout-removing the masks then performs the layout.
  
  For the masking experiments, we begin by a toy problem where a sequence of numbers is provided to the transformer. A specific number within this sequence is masked according to a particular 
  order, and experimented whether transformers can understand that pattern. 
  We then continue to apply the masking strategy to synthetic netlists that we created after several discussions with subject matter experts. In this dataset the transistors are grouped based on 
  having interconnected components of similar types through source-drain or bulk sharing. 
  This appeared to us as a natural first step towards full circuit layout, as discovered groups of transistors can (but not need to) be placed as larger placement components.
  
  \subsubsection{Toy Problem}
  We designed a toy problem involving masking a number in a grid and having the transformer predict the value of the 4th item in the row preceding the masked row. 
  The objective of this toy problem was to evaluate whether GPT-2 can reason in a 2D space and understand spatial relationships among tokens in a 2D grid, specifically the relationship between the masked 
  token and a spatial pattern relative to it. 
  A sample input sequences is given in Fig.~\ref{fig:toy_gpt}:
  \begin{figure}[!htbp]
  \centering
  \begin{tcolorbox}[
    colback=gray!5, colframe=black!50,
    listing only,
    listing options={basicstyle=\ttfamily\tiny, breaklines=true},
    width=\columnwidth, 
  ]
  ‘0,0:9;0,1:1;0,2:0;0,3:0;0,4:4;1,0:5;1,1:9;1,2:3;1,3:3;1,’
  \end{tcolorbox}
  \caption{Sample training data for the toy problem with GPT2: masking a number in a grid and having the transformer predict the value of the 4th item in the row preceding the masked row.}
  \label{fig:toy_gpt}
  \end{figure}
  
  This can be interpreted to mean that the item at coordinates (0,0) is 9, (0,1) is 1, (0,2) is 0, (0,3) is 0, and so on.
  Our results show that GPT2 was unable to solve this problem, and loss did not decrease below ~2.3. Given that GPT2 was unable to output the correct answers, we created some simplified toy problems to 
  target what exactly GPT2 was struggling with. First, we created a training dataset of sequences that were 10 integers long, with each integer being between 0 and 5. An index was randomly selected to mask, 
  and the model was trained to output the index of the masked integer. A FlaxGPT2 model with $number of layer=2$ and $number of head=2$ was able to quickly reach $100\%$ accuracy on this problem.
  Next, we attempted the problem of predicting the index N before the mask. For example, for N=7, we would have the sequences and labels provided in Fig~\ref{fig:toy_gpt_1D}:  
  \begin{figure}[!htbp]
  \centering
  \begin{tcolorbox}[
    colback=gray!5, colframe=black!50,
    listing only,
    listing options={basicstyle=\ttfamily\tiny, breaklines=true},
    width=\columnwidth, 
  ]
  Sequence 1: 2,6,8,5,6,9,9,4,100,2,9,3,0,3,5,3,4,7,2,7
  
  Label 1: 6
  
  Sequence 2: 5,4,5,4,5,8,7,6,100,4,8,2,5,6,6,2,8,5,2,4
  
  Label 2: 4
  
  Sequence 3: 8,6,7,3,6,4,4,3,5,100,0,8,6,1,4,9,3,0,5,7
  
  Label 3: 7
  \end{tcolorbox}
  \caption{Sample training data for the 1-D version of the toy problem with GPT2: predicting the index N before the mask, N is 7 in this example.}
  \label{fig:toy_gpt_1D}
  \end{figure}
  
  This is a 1-Dimensional version of the grid task that we described as our original toy problem.
  GPT2 struggled with solving this problem. For N=1, we reached a maximum of $50\%$ accuracy. For N=4 and N=7, the accuracies were much lower. To attempt to encourage the model to learn spatial 
  relationships, we added a convolutional layer and evaluated the model using the following steps:
  \begin{itemize}
    \item Input text tokenized
    \item Tokenized input converted to GPT2 embeddings
    \item Position embedding added with retrained GPT2 embeddings
    \item Combined embeddings passed thru 1D convolution
    \item Output of convolution layer is fed into GPT2 transformer blocks
    \item Final hidden state passed thru projection to predict next token
  \end{itemize}
  
  With the convolutional model, the training converged very quickly and the model achieved $100\%$ accuracy. The higher the N value, the longer the model took to train to $100\%$ accuracy. 
  However, for all N <=7, the models all converged in 2 epochs or fewer.
  
  As a final test, we generated 10-integer long sequences with integers between 0-9 with a random number of randomly masked values. The model was trained to count the number of masked tokens. 
  A sample training data is provided in Fig.~\ref{fig:toy_gpt_random}.
  \begin{figure}[!htbp]
  \centering
  \begin{tcolorbox}[
    colback=gray!5, colframe=black!50,
    listing only,
    listing options={basicstyle=\ttfamily\tiny, breaklines=true},
    width=\columnwidth,
  ]
  Sequence 1: 100,4,100,100,100,0,5,100,100,100
  
  Label 1: 7
  
  Sequence 2: 100,100,100,100,100,100,100,100,2,100
  
  Label 2: 9
  
  Sequence 3: 1,1,100,3,3,100,100,100,100,100
  
  Label 3: 6
  \end{tcolorbox}
  \caption{Sample training data for the randomly masked token experiments: 10-integer long sequences with integers between 0-9 with a random number of randomly masked values. }
  \label{fig:toy_gpt_random}
  \end{figure}
  
  GPT2 without the convolutional layer was able to quickly solve this exercise.
  To summerize our observetaions from the above experiments:  GPT-2 struggled to learn complex 2D spatial relationships. It performed slightly better with 1D relationships. 
  However, incorporating convolutional layers significantly improved performance, ensuring that spatial relationships are properly learned.
  \subsubsection{Netlist Subgroup Masking} 
  
  A critical hurdle to applying large language models for analog circuit layout is the lack of availability of large datasets of circuit schematics.
  To address this challenge, we intially worked to generate synthetic analog circuit schematics and layout information to train and fine-tune models.
  In the application of Large Language Models for analog circuit design is how to represent circuit schematics in a meaningful way.
  
  We made an attempt to make a dataset that the large language model can extract relevant information from while ignoring unimportant noise and variability in inputs. 
  We began by generating realistic SPICE style netlists that display relational properties that are demonstrated across analog circuits (e.g., drain-source sharing).
  Each generated circuit is comprised of a sequence of transistor-like blocks, each of which is defined by four identifiers corresponding to terminal connections (e.g., gate, source, drain, and bulk), 
  the component type (e.g., pshort, nshort), and both a group and subgroup identifier. 
  The structure of each component generated in a netlist appears as: $ID_{\text{drain}}$, $ID_{\text{gate}}$, $ID_{\text{source}}$, $ID_{\text{bulk}}$, $\text{type}(ID_{\text{group}}, ID_{\text{subgroup}})$
  
  To simulate realistic circuit structure, we generate transistor-like components in chains that share specific pin values.
  Block group and subgroup membership of each component is defined according to heuristics that describe sections of transistors that can be combined and/or placed in the same region. 
  Subgroup membership is permitted, but not required, between circuit components (e.g., Mosfet transistors) that share drain-source connections ($ID_{drain}, ID_{source}$) .
  Larger group membership combines transistors with common body tie or $ID_{bulk}$ connection. $ID_{gate}$ may vary between components in the same group or subgroup.
  
  \begin{figure}[!htbp]
    \centering
    \includegraphics[width=\columnwidth]{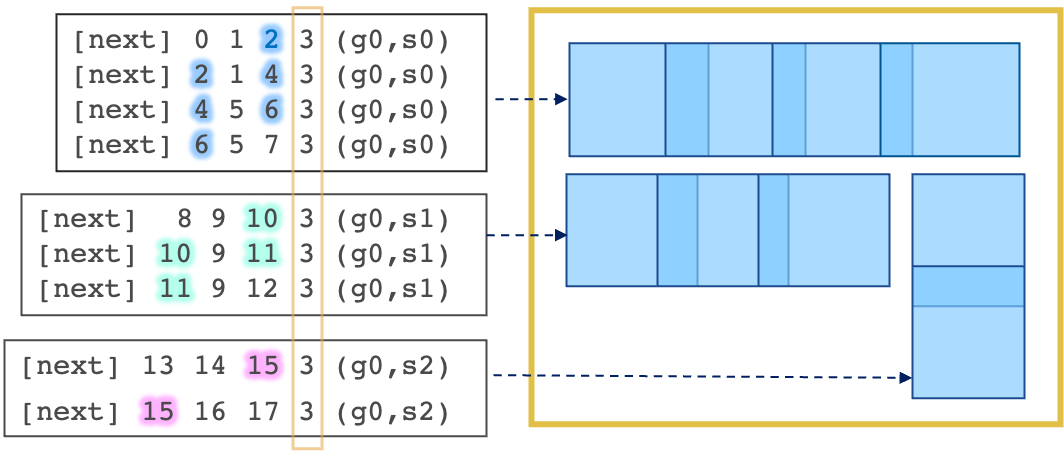}
    \caption{An example and illustration of a synthetic netlist group (g0) consisting of 3 subgroups (s0, s1, s2). Each component is defined after an initial [next] token by its terminal connections and group/subgroup membership.}\label{fig:synthetic_netlist}
  \end{figure}
  
  Figure \ref{fig:synthetic_netlist} shows an example of a synthetically generated netlist group, g0.
  
  We began with a sub-problem of circuit layout placement: \textit{What transistors should be grouped together via source-drain sharing?} \textit{And what shared groupings should be grouped with the same body tie?} 
  By masking groups and subgroups in our synthetic circuit representations, we assess the ability large language models to learn this sub-task critical for circuit layout.
  
  We find that the accuracy of a T5 transformer model in unmasking component groupings decreases as the number of components masked is increased. 
  This is unsurprising, as the complexity of the unmasking task increases with the masking of additional group and subgroup tokens. 
  
  \begin{table}[ht]
  \caption{T5 accuracy unmasking synthetic netlist groups.}
  \centering
  \begin{tabular}{|c|c|}
  \hline
  \textbf{\# groups masked} & \textbf{Accuracy} \\
  \hline
  1 & 85.95 \\
  \hline
  2 & 73.67 \\
  \hline
  3 & 65.96 \\
  \hline
  4 & 56.89 \\
  \hline
  \end{tabular}
  \label{table:t5grouping}
  \end{table}

  \subsubsection{Masking for Layout Generation: }\label{sec:grid_t5}
  In an effort to address the layout and placement challenges in analog circuit design, we generated a synthetic dataset designed for both in-context learning and model fine-tuning. 
  The goal of this task was to place transistor groups represented as rectangles onto a discrete grid in a way that satisfies three primary constraints: symmetry with respect to the y-axis, non-overlapping placement, and space minimization. 
  The dataset comprises 10,000 samples and is split into training, validation, and test sets. Each sample corresponds to a unique layout scenario, in which multiple rectangles must be placed according to specific rules. 
  Each transistor is modeled as a group of contiguous $1 \times 1$ squares, and the layout is described using the following schema:
  
  \begin{itemize}
      \item \texttt{sb\{i\}}: A group identifier, where $i$ is the transistor index.
      \item \texttt{P/N}: The type of transistor, either P-type or N-type.
      \item \texttt{(x, y)}: The starting coordinates of the rectangle on the grid.
      \item \texttt{m}: The number of $1 \times 1$ squares comprising the rectangle.
      \item \texttt{b}: A boolean indicating whether the rectangle is rotated by $90^\circ$.
  \end{itemize}
  
  Each sample is therefore represented as a list of placement instructions in the form:
  
  \begin{verbatim}
  sb{i}(T, (x, y), m, b)
  \end{verbatim}
  
  where $T \in \{P, N\}$, $(x, y)$ denotes the top-left coordinate of the rectangle, $m$ is its length (depending on orientation), and $b$ indicates rotation.
  This representation facilitates structured reasoning and pattern extraction by language models, providing a controlled and interpretable environment for experimentation on analog layout generation.
  
  We then train a T5 small transformer to predict the masked transformer in the sequence such that symmetry is preserved. To account for non-overlapping criteria, a reguralization parameter is 
  added such that the loss would increase drastically if there is an overlap between components. A sample prompt that is used to train th T5 small model is given in Fig.~\ref{fig:layout_masking}:
  \begin{figure}[!htbp]
  \centering
  \begin{tcolorbox}[
    colback=gray!5, colframe=black!50,
    listing only,
    listing options={basicstyle=\ttfamily\tiny, breaklines=true},
    width=\columnwidth, 
    title=Model Prompt
  ]
  fixed components: ['sb1(P, (0, 16), 4, False)', MASK ,
  'sb3(N, (4, 16), 3, False)', ...'],
  
  next component:'P','4', 'sb2(P, (16, 16), 4, False)'
  \end{tcolorbox}
  \caption{Sample training data for the masking for layout generation experiment: one transistor is masked and the model is trained to output the masked transistor}
  \label{fig:layout_masking}
  \end{figure}
  
\begin{figure}[!htbp]
    \centering
    \begin{subfigure}[b]{0.49\columnwidth}
        \centering
        \includegraphics[width=\columnwidth]{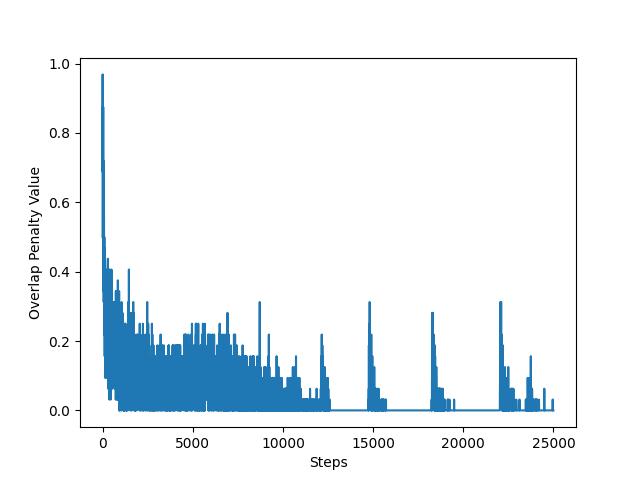}
    \end{subfigure}
    \hfill
    \begin{subfigure}[b]{0.49\columnwidth}
        \centering
        \includegraphics[width=\columnwidth]{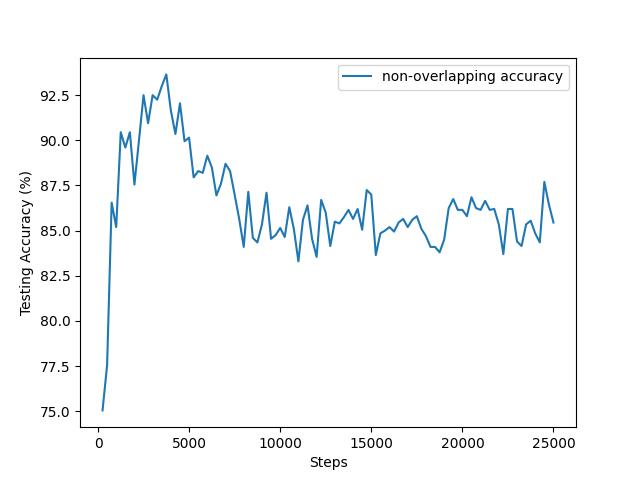}
    \end{subfigure}
    \caption{Left: Overlapping Penalty, Right:Testing Accuracy}
    \label{fig:t5_train_grid}
  \end{figure}

  Fig~\ref{fig:t5_train_grid} shows the results of our trained model on test data. During training, the overlap penalty is decreasing, meaning that the model is successfully learning how to avoid overlaps 
  without explicitly being advised to in the prompt. Although the model is not able to generate the exact same transistor locations which is very expected as the transistor locations where 
  generated randomly in making the synthetic dataset, it is very useful in generating locations 
  on the grid which are not overlapping with already fixed transistors (non-overlapping accuracy of about 87\%).
  
\begin{figure}[!htbp]
    \centering
    \begin{subfigure}[b]{0.49\columnwidth}
        \centering
        \includegraphics[trim=0 40 0 0,clip,width=\columnwidth]{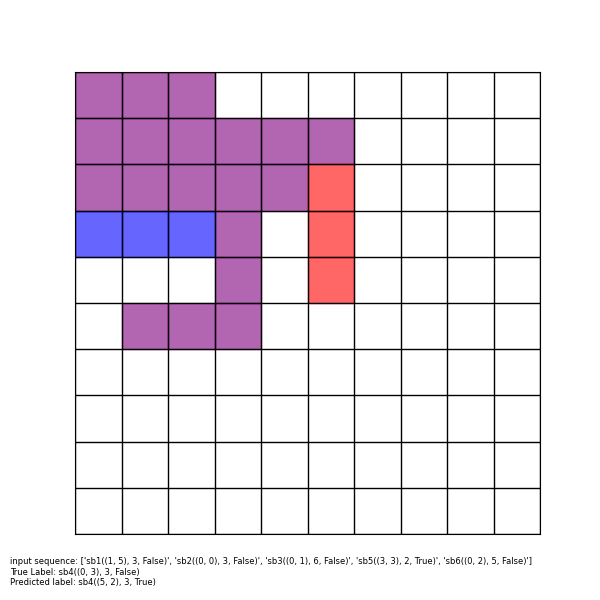}
    \end{subfigure}
    \hfill
    \begin{subfigure}[b]{0.49\columnwidth}
        \centering
        \includegraphics[trim=0 40 0 0,clip,width=\columnwidth]{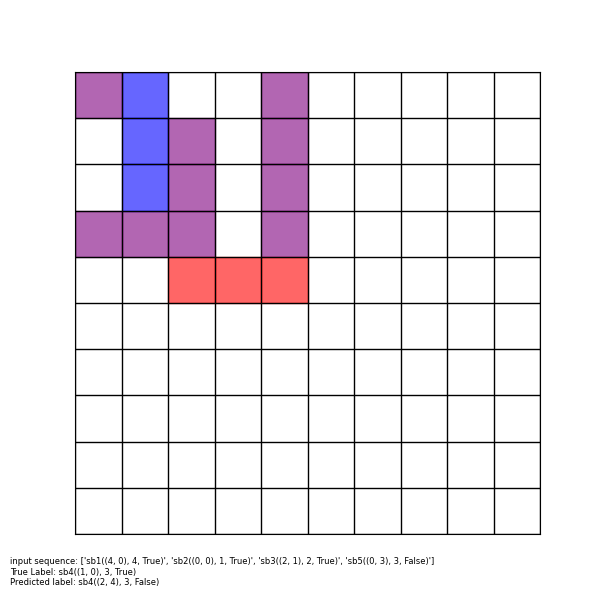}
    \end{subfigure}
    \caption{Sample transistor locations on a $20 \times 20$ grid. Already placed transistors are in purple, the true label from the test data is in blue, and the predicted one is in red.}
    \label{fig:grids_t5}
  \end{figure}

  Fig~\ref{fig:grids_t5} shows sample test results from our trained model.Already placed transistors are in purple, the true label from the test data is in blue, and the predicted one is in red.
  This shows that the model is generating transistors that are not overlapping with the ones that are already placed, but they are not the same size, or at the same location as the ground truth data.
  As the data was generated randomly, this outcome is fully expected, which led to our next investigation for making a more realistic synthetic data, and finetuning with better prefixes.
  
  \subsection{Sequential Placement \(Seq2Seq\) Layout Design: }
  Building on the spatial understanding gained from the masking stage, we fine-tuned a Mistral7B model to perform sequence-to-sequence transistor placement. 
  In this experiment, the model received as input the spatial features of an initial transistor and was tasked to iteratively generate the subsequent transistors. 
  The objective was to ensure that critical layout characteristics—such as symmetry, spacing constraints, and relative positioning—were preserved across the sequence. 
  This formulation allowed the model to reason about placement dependencies in a stepwise manner which is elaborated on in the following sections.
  
  \subsubsection{Fine-tuning Mistral7B with Synthetic Data v1:}
  In this experiment, we utilized a synthetic dataset dicuseed in section~\ref{sec:grid_t5} to fine-tune a Mistral7B model. The objective was to enable the model to generate the next transistor 
  location on a grid. To guide the language model (LLM) towards the task, we added a prefix that specified the task requirements, emphasizing symmetry and non-overlapping placement, and explained 
  the significance of each number in the dataset. Additionally, we provided examples demonstrating how transistors are placed on the grid. 
  Subsequently, we queried the model to determine the optimal placement for the next transistor. We incorporated prefix tuning to provide the model with some context about 
  the task. Full experiment details including the full prompt are given in Appendix. 
  
\begin{figure}[!htbp]
    \centering
    \begin{subfigure}[b]{0.3\textwidth}
        \centering
        \includegraphics[width=\textwidth]{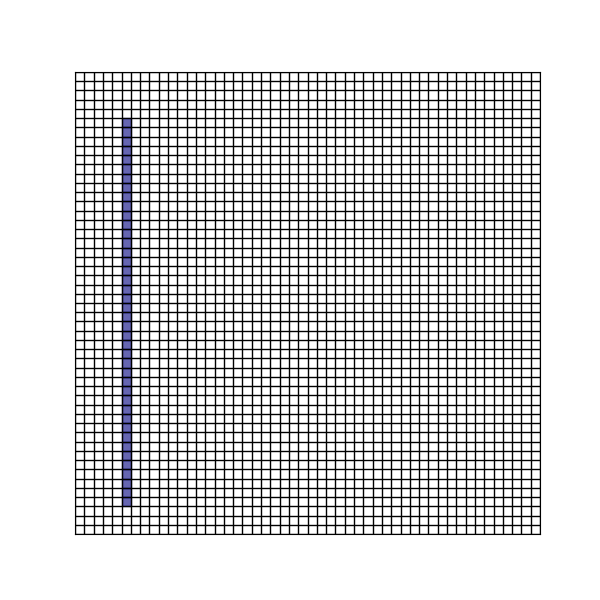}
    \end{subfigure}
    \hfill
    \begin{subfigure}[b]{0.3\textwidth}
      \centering
      \includegraphics[width=\textwidth]{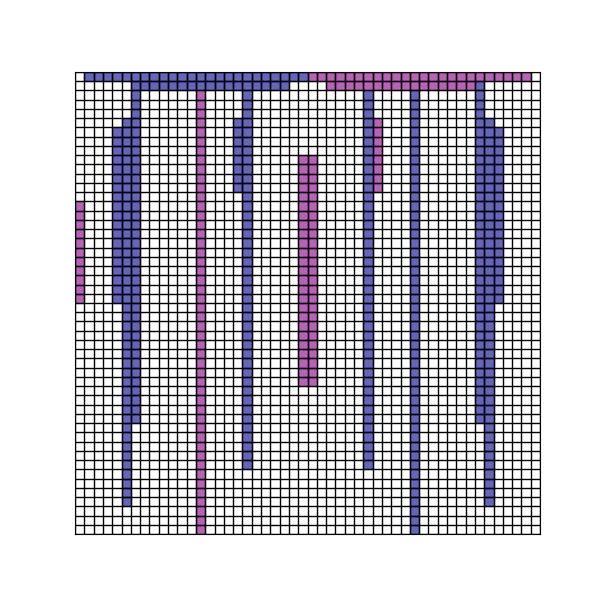}
    \end{subfigure}
    \hfill
    \begin{subfigure}[b]{0.3\textwidth}
      \centering
      \includegraphics[width=\textwidth]{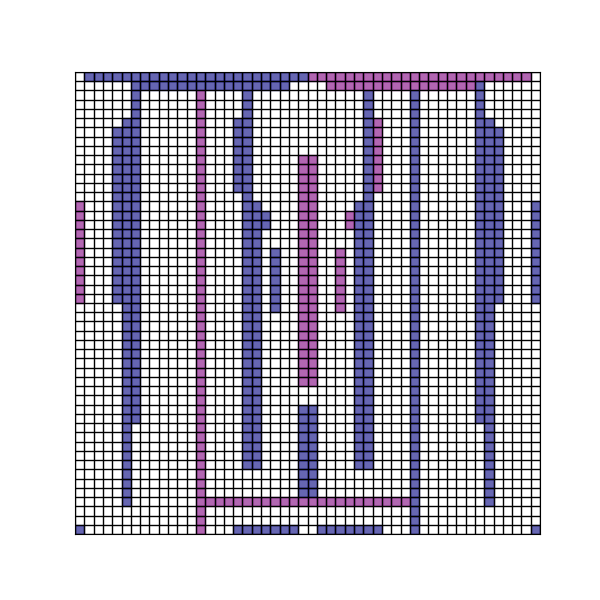}
    \end{subfigure}
    \caption{Sample transistor locations on a $50 \times 50$ grid. left: first transistor is being placed at a random location. 
    middle: the 18th transistor is placed such that the symmetry, and non-overlapping criteria is preserved. right: last transistor in the sequence is placed such that constrains are met}
    \label{fig:grids_finetune}
  \end{figure}

  The fine-tuning process was conducted using the PEFT package and open-source LoRA parameters for $1$ epoch. Detailed hyperparameters used in the experiment are available in our code repository. 
  We monitored the training, validation, and testing loss throughout the process. Upon inferencing with the fine-tuned model, we observed that it successfully generated layouts that maintained symmetry 
  and avoided overlap as seen in Fig~\ref{fig:grids_finetune}. The results indicate that the fine-tuned Mistral7B model is effective in generating transistor placements that adhere to the specified constraints, 
  demonstrating its potential utility in automated layout generation tasks.
  
  The results indicate that the fine-tuned Mistral7B model is effective in generating transistor placements that adhere to the specified constraints, 
  demonstrating its potential utility in automated layout generation tasks.
  While the fine-tuned Mistral7B model is effective in generating layouts with certain constraints, it is not the most practical solution for real-world applications. 
  In real-world settings, system components such as width, height, and types of transistors are fixed and predefined by the user. Therefore, to be truly useful in practical scenarios, 
  a model must be capable of placing components one by one, given these fixed parameters. This necessitates the development of a more advanced model that can accommodate fixed widths, heights, 
  and types of transistors as input parameters, ensuring accurate and feasible layout generation which led us to further optimize our synthetic dataset.
  \subsubsection{Fine-tuning Mistral7B with Synthetic Data v2:}
  
  In the subsequent phase of our research, we refined our approach to better align with real-world requirements. We revised the model prompt to allow users to specify the name of the subgroup, 
  the type of transistor, and its width and height, and then query the model for the centerpoints of the transistors. This adjustment ensures that the generated layouts adhere to predefined dimensions 
  and types, facilitating practical application. To achieve this, we constructed a dataset where the transistor placements were symmetrical and non-overlapping, ensuring that the model 
  could learn these constraints effectively. The revised prompt was designed to focus solely on generating the centerpoints of the transistors, given the fixed parameters provided by the user. 
  A description of the full prompt is given in the Appendix.
  
\begin{figure}[!htbp]
    \centering
    \begin{subfigure}[b]{0.49\columnwidth}
      \centering
      \includegraphics[width=\columnwidth]{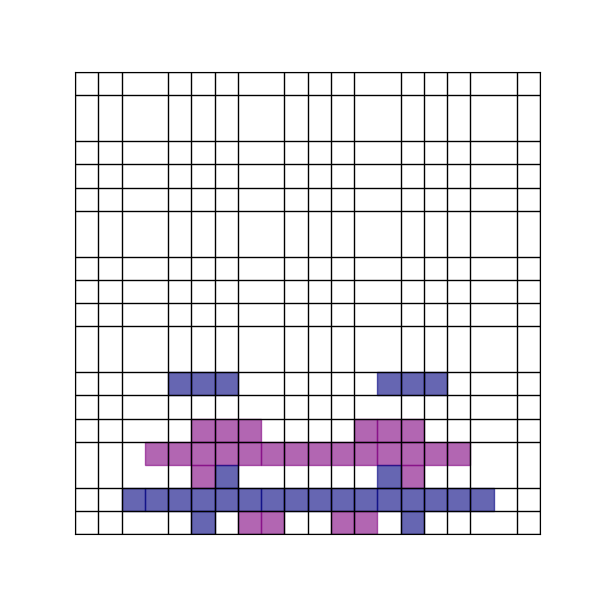}
    \end{subfigure}
    \hfill
    \begin{subfigure}[b]{0.49\columnwidth}
      \centering
      \includegraphics[width=\columnwidth]{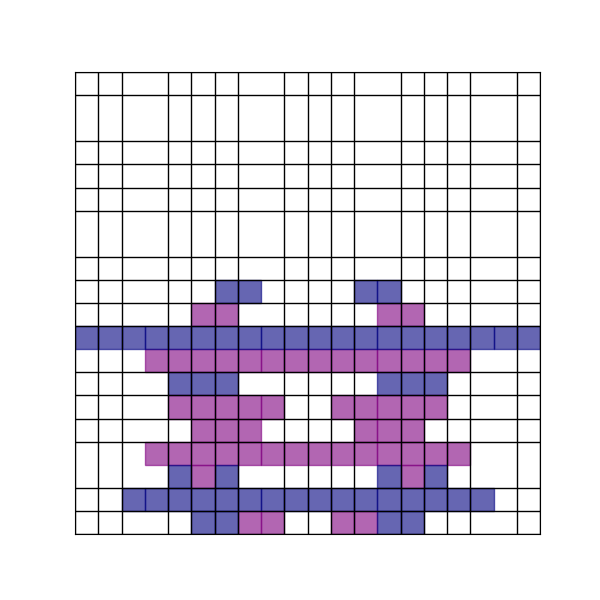}
    \end{subfigure}
    \caption{Sample transistor groups that the model is prompted on. left: the 18th transistor is placed such that the symmetry, non-overlapping, and space minimization criteria is preserved. 
    right: last transistor in the sequence is placed such that constrains are met}
    \label{fig:centerpoints}
  \end{figure}

  This approach allowed us to maintain the integrity of the layout while accommodating the fixed dimensions and types specified by the user. The results from this refined model demonstrated 
  improved practicality and usability in real-world scenarios, where precise component placement is crucial. Fig~\ref{fig:centerpoints} shows a sample synthetic netlist that the model is prompted to 
  solve the placement task for, which shows that the fine-tuned model is capable of placing transistors such that the conditions mentioned in the prompt are met. 
  
  This approach allowed us to maintain the integrity of the layout while accommodating the fixed dimensions and types specified by the user. 
  The results from this refined model demonstrated improved practicality and usability in real-world scenarios, where precise component placement is crucial.
  \subsubsection{Mistral7B Evaluation on Real-world Circuits:}
  
\begin{figure}[!htbp]
    \centering
    \begin{subfigure}[b]{\columnwidth}
        \centering
        \includegraphics[width=\columnwidth]{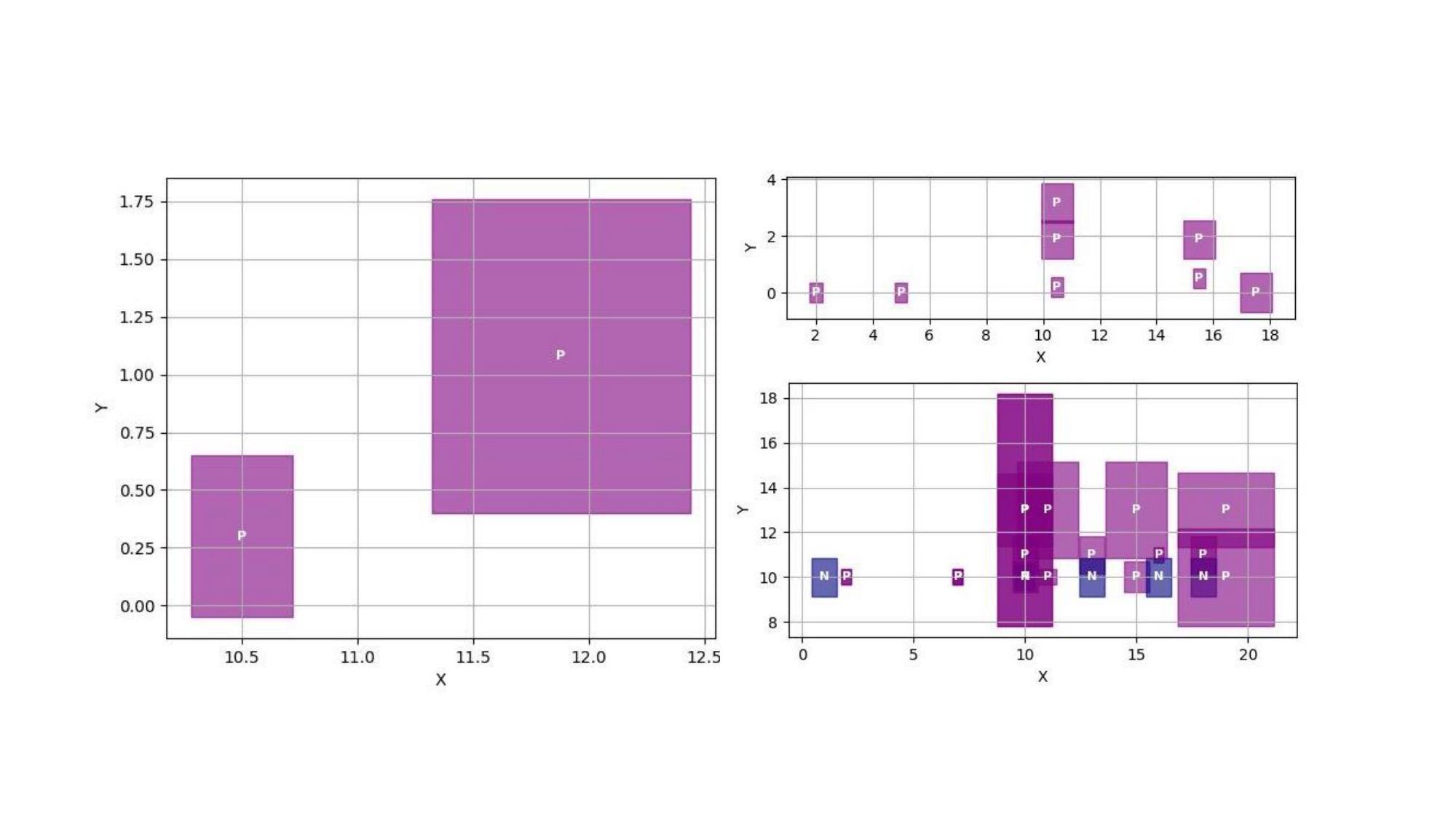}
    \end{subfigure}
    \caption{Mistral-7B placement results using real transistors in a netlist: the leftmost netlist shows that when the number of components are low, the model works much better compared to the netlists on the right 
    where there are more transistors to place. A small overlap is seen in the upper-right netlist, but symmetry and space minimization is not satisfied in any of the layouts.}
    \label{fig:real}
  \end{figure}

  In the subsequent step of our research, we sought guidance from subject matter experts to gain insights into the characteristics 
  of real-world netlists, including dimensions, wiring placements, buffers, and other relevant parameters. This expert input was 
  crucial for further refining our approach to better align with practical applications. The Fig~\ref{fig:real} illustrates the results 
  of applying our fine-tuned model to real-world netlists, which exhibit a slightly different format compared to our synthetic dataset. As depicted in the figure, while the model successfully minimizes space utilization, it introduces significant overlaps and fails to preserve symmetry.
  These findings underscore the challenges associated with adapting our model to real-world scenarios. The discrepancies highlight 
  the need for additional refinements to ensure that the generated layouts not only minimize space but also adhere to constraints such 
  as non-overlapping placements and symmetry preservation.
  
  \subsection{Joint \(All-at-Once\) Placement Layout Design: }
  In the final stage, we transitioned from sequential prediction to joint placement generation. 
  Instead of producing placements one at a time, the model was prompted with all transistor coordinates simultaneously and tasked to output the complete layout in a single pass. 
  This approach, supported by a larger model and a more context-rich prompt, enabled the model to reason about global structure and spatial coherence across all transistors. 
  The joint formulation also mitigated error propagation inherent in sequential models and yielded more globally consistent layouts which is discussed below.
  
  \subsubsection{Fine-tuning GPT20B with Synthetic Data v3: }
  When evaluating the fine-tuned Mistral-7B model on real netlists, we observed that perfect symmetry could not always be preserved due to practical design constraints. 
  After consulting with subject matter experts and confirming that such asymmetries are common in real-world layouts, we revised our synthetic dataset to better reflect these 
  conditions. Specifically, we introduced controlled asymmetry by allowing symmetry violations in approximately $20\%$ of the training samples.
  
  \begin{figure}[!htbp]
  \centering
  \begin{tcolorbox}[
    colback=gray!5, colframe=black!50,
    listing only,
    listing options={basicstyle=\ttfamily\tiny, breaklines=true},
    width=\columnwidth, 
  ]
  You are an expert in analog circuit design. The task is to place transistors on a layout.
  Each transistor is described by its id (e.g., MN0), type (e.g., P/N or technology-specific strings
  such as pthick/narrow), width (x), and height (y). Your goal is to assign placement coordinates
  ($c_x$, $c_y$) for each transistor such that:
  \begin{itemize}
  \item No two transistors overlap.
  \item Transistors of the same type (P or N, or equivalent type string) are grouped together as much as possible.
  \item The layout is as compact as possible, minimizing total area.
  \item Symmetry with respect to the y-axis should be preserved when feasible, but it is not mandatory.
  \item The output must preserve the original list order and return the same fields (id, type, x, y)
  augmented with optimized $c_x$ and $c_y$.
  \end{itemize}
  Think carefully about placement before outputting. Return the final placement as a JSON list
  of transistor objects.
  
  Only output the JSON list of transistors with fields id, type, x, y, $c_x$, $c_y$. Do not include any extra text or explanation.
  
  Now here is the input:
  [
    {"x": 3.92, "y": 8.36, "id": "sb2", "type": "P"},
    {"x": 3.14, "y": 8.36, "id": "sb1", "type": "P"},
    {"x": 2.75, "y": 8.36, "id": "sb0", "type": "P"}
  ]
  \end{tcolorbox}
  \caption{Sample prompt for the finetuning for all-at-once layout generation experiment using data v3: all the transistor information is given in the prompt, 
  the model is finetuned to place transistors given the height and width. It will output the centerpoints accounting for the constraints given in the prompt}
  \label{fig:prompt_v3}
  \end{figure}

  To further improve robustness under these realistic conditions, we fine-tuned a larger GPT-OSS-20B model. 
  This model was optimized to produce placements that maintain non-overlapping transistors, minimize layout area, and preserve symmetry whenever feasible.
  In this phase, we also moved away from grid-based representations, as they inherently constrained the (x, y) coordinates to integer values—an assumption that does 
  not hold in practical circuit layouts. Instead, we adopted a continuous coordinate formulation that allows finer spatial resolution and better alignment with real-world transistor geometries. 
  The sample data used for training is shown in Fig.~\ref{fig:tr_data_gpt}, which illustrates a more realistic representation of transistor placements observed in real circuits (shown in Fig.~\ref{fig:real}).
  The prompt used during fine-tuning is shown below:
  \begin{figure}[!htbp]
    \centering
    \begin{subfigure}[b]{0.98\columnwidth}
        \centering
        \includegraphics[width=0.98\columnwidth]{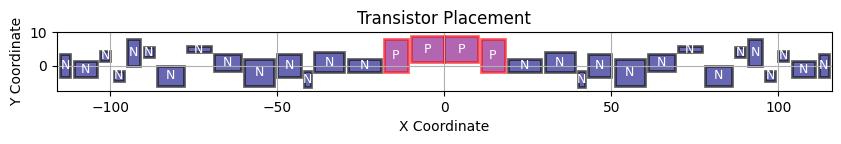}
    \end{subfigure}
    \hfill
    \begin{subfigure}[b]{0.98\columnwidth}
      \centering
      \includegraphics[width=0.98\columnwidth]{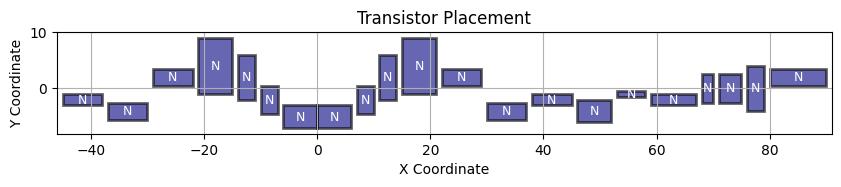}
    \end{subfigure}
    \caption{Sample transistor groups that the GPT20B model is prompted on.}
    \label{fig:tr_data_gpt}
  \end{figure}
  
  \subsubsection{Parsing the LLM Output: }
  Parsing the output of large language models for layout generation is non-trivial. To fully close the loop and use the model for automated transistor placement, 
  it is critical to extract exactly the fields we need—specifically id, type, x, y, $c_x$, and $c_y$—from potentially verbose model responses. 
  We explored two parsing strategies: (i) using the regex Python package to identify structured patterns, and (ii) using a secondary LLM as a “judge” to extract 
  and validate the JSON output. A sample model output is provied in Fig~\ref{fig:output_promptv3}.
  
  Based on the sample model outputs, extracting the center coordinates ($c_{x}$, $c_{y}$) for each transistor directly from the raw text is often infeasible using simple 
  regex-based parsing. To address this, we adopted an agentic LLM approach, where a secondary model (e.g., Zephyr-7B-Beta) was prompted to extract and structure the centerpoints 
  from the primary model output. The parsing component of this pipeline remains an active area of research, particularly regarding whether fine-tuning dedicated LLMs can further 
  improve the accuracy and reliability of structured data extraction.
  
  
  \subsubsection{Finetuned GPT-oss-20B Model Evaluation on Real-world Circuits: }
\begin{figure}[!htbp]
    \centering
    \begin{subfigure}[b]{0.49\columnwidth}
        \centering
        \includegraphics[width=0.4\columnwidth]{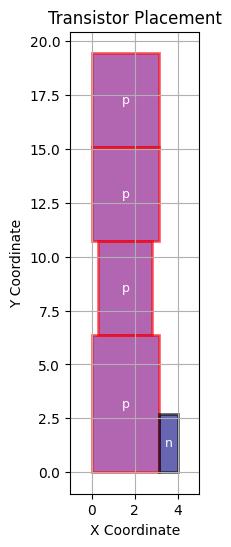}
    \end{subfigure}
    \begin{subfigure}[b]{0.49\columnwidth}
      \centering
      \includegraphics[width=0.4\columnwidth]{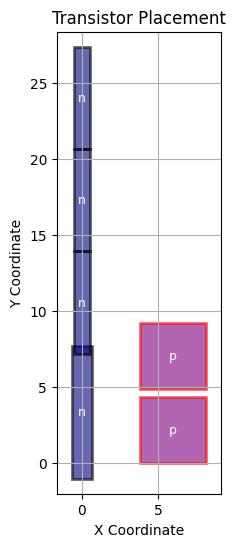}
    \end{subfigure}
    \begin{subfigure}[b]{\columnwidth}
      \centering
      \includegraphics[width=0.3\columnwidth]{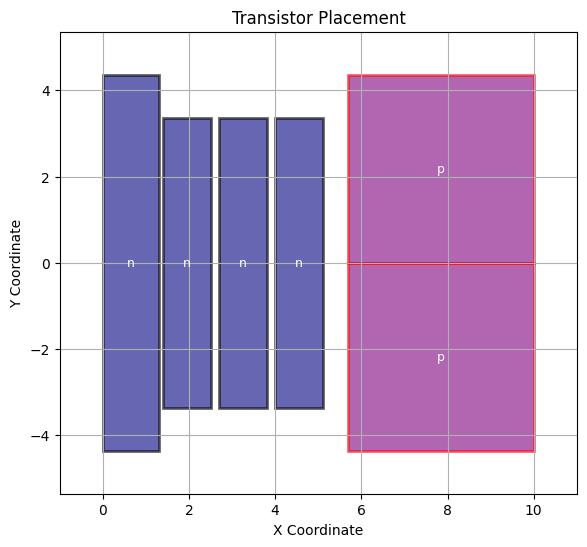}
    \end{subfigure}
    \hfill
    \caption{GPT-oss-20B placement results using real transistors in a netlist: these three samples show that the model is successfully minimizing 
    the space while perserving the non-overlapping criteria, as well as attending to symmetry as much as possible. However, a very tiny overlap is visible in the upper-right netlist}
    \label{fig:real-gpt20}
  \end{figure}

  We subsequently repeated the transistor placement experiments on the real industry data using the larger GPT-OSS-20B model, following the same experimental setup as with the Mistral-7B model. 
  The same prompt prefix and structure were employed, while real netlist data were provided as input samples. 
  The model was queried to generate the corresponding placement center coordinates for each transistor.
  Representative results are illustrated in Fig.~\ref{fig:real-gpt20}, which demonstrate that the generated placements exhibit no overlap among transistors and maintain consistent 
  grouping and near-symmetric structures. These outcomes suggest that the GPT-OSS-20B model, when combined with a more realistic and diversified dataset, 
  is capable of effectively learning and replicating the spatial organization principles underlying transistor placement.
  Overall, the results indicate a strong potential for large language models to perform layout reasoning and geometric optimization in analog design tasks. 
  However, challenges remain—particularly in reliably parsing model outputs, mitigating hallucinated placements, and improving robustness under unseen design topologies. 
  Addressing these limitations represents an important direction for future work.
  
  \section{Conclusion and Future Work}
  
  In this paper, we explored the application of transformer architectures with variable sizes for analogue circuit design, addressing real-world engineering challenges.
  We began by utilizing smaller models to solve a toy problem of finding masked values in sequences and progressively advanced to generating masked circuit components using models like T5. 
  Our approach was further extended by fine-tuning Mistral7B, and GPT-oss-20B LLMs to determine the x,y positions of transistors in compliance with DRC rules. 
  Through synthetic data, we demonstrated that appropriate prompting could generate designs meeting criteria such as symmetry, space minimization, and non-overlapping specifically on real-world netlists. 
  Smaller models fine-tuned on immature synthetic data faced challenges when evaluated on real-world data due to format complexities; 
  these issues were mitigated by adopting a more advanced model architecture and SME-driven data representations.
  
  Aside from more robust model output parsing strategies, future work focuses on developing a hypergraph-based representation to capture structural information, enabling a multimodal fine-tuning approach that 
  integrates both topological and symbolic data. We are also leveraging graph representations to improve transistor grouping and exploring LLaMA 4 for multimodal tasks, 
  which shows promising potential. By pursuing these directions, we aim to enhance the capabilities of transformer architectures for analog circuit design, 
  ultimately supporting more efficient and innovative engineering solutions.
  
  \clearpage
  \section{Appendix}\label{app:drc}
  
  \subsection{Design Rule Check: }\label{drc}
  Design rules are a vitally important aspect of layout design. Design Rule Checking (DRC) is the process of making sure you follow these rules, almost always using software.
  These rules are supplied by the foundry for each process. Breaking these rules can lead to an issue of yield, meaning that a percentage of your devices may be broken simply by the stresses of the manufacturing process.
  
  Foundry is used to refer to a company/facility that manufactures circuits, usually integrated circuits (IC). Typically, this refers to the company, but the facility and machinery used are also called the foundry. 
  The methodology that the foundry will use to create the circuit is called a "process" of that foundry. 
  
  A few examples of important Design Rules include basic checks, like spacing between certain materials, the widths of said materials, and minimum areas.
  
  In addition to DRC, there is LVS, meaning "Layout vs Source". This is another check much like DRC. it is done via software.
  
  LVS gives you feedback on how closely a completed layout matches the "source". This source is, in anolog layout, a netlist (often generated from a circuit diagram).
  The tool takes the completed layout and, programatically, generates a netlist that represents it. Then, it simply checks for differences between the Source and Layout netlist.
  
  \subsection{Circuit Design Challenges: }\label{app:circuit_challenges}
  
  Circuit layout design can be broken down into 3 stages, each with their own set of challenges discussed in Appendix~\ref{app:circuit_challenges}.
  
  First is floorplanning - essentially, choosing relative placements for all devices. There are many aspects to consider. Matching is ensuring that certain devices are placed in a similar environment, such as sorrounded by similar devices.
  You also want to make sure that sensitive circuits are made symmetrical. You also have some relatively simple design rules to consider, such as widht and spacing rules.
  some more complex design rules apply here as well, such as DFM rules. These are rules that are designed to improve yield and simplify manufacturing when possible, and sometimes can be bent if needed.
  One of the most important non-DRC considerations at this stage is parasitics. These are devices occur in a layout unintentionally, such as a diode appearing across the boundaries of certain structures. This specific parasitic can lead to an issue called Latchup.
  
  \begin{figure}[!htbp]
    \centering
    \includegraphics[width=\columnwidth]{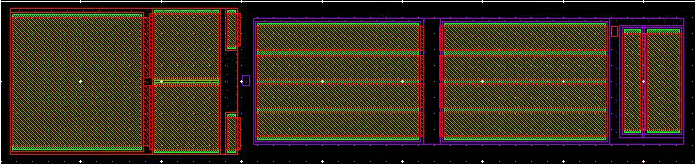}
    \caption{an example layout after the floorplanning stage, but prior to the routing stage.}\label{fig:LayoutI}
  \end{figure}

  Second, you connect the devices. This is often the quickest step, but requires attention to make the last step go smoothly. Here, you place down metal routes to connect each device as they are defined by the netlist.
  Special care must be taken to follow metal width and spacing rules, but also to ensure no problematic shorts (where two metal routes connect, where they should not) or opens (where two metal routes do not connect, where they should) occur.
  There are still some more complex considerations at this stage. Antenna is an effect where a long metal route can pick up an electric charge during manufacture, causing a disasturous discharge when later connected to a sensitive structure.
  In addition, density of a metal across a chip can cause the material properties of the chip to change. Parisitics come into play here as well, introducing parasitic resistance and capacitance, which may need to be minimized.
  
  \begin{figure}[!htbp]
    \centering
    \includegraphics[width=\columnwidth]{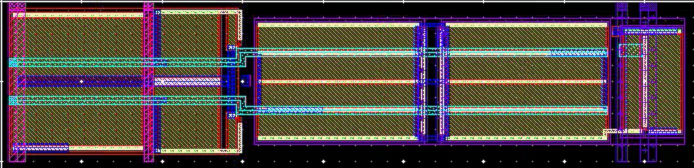}
    \caption{an example layout after routing. This layout also passes both DRC and LVS, meaning this is a completed layout.}\label{fig:LayoutII}
  \end{figure}
  
  Lastly, you run DRC and LVS checks, and modify the layout as needed. The difficulty of this stage directly depends on how well you followed the rules and the netlist on your first pass. failure at this stage means revisiting previous steps of the process.
  
  \subsection{Fine-tuning Mistral7B with Synthetic Data v1 Experiment Details: }
  
  Fig~\ref{fig:layout_finetune} shows a sample prompt that the model was fine-tuned with:
  
  \begin{figure}[!htbp]
  \centering
  \begin{tcolorbox}[
    colback=gray!5, colframe=black!50,
  listing only,
  listing options={basicstyle=\ttfamily\tiny, breaklines=true},
  width=\columnwidth,
  ]
  "The task is to place groups of squares in a way that they don't overlap, they are symmetrical with respect to the y axis, and they are clustered and grouped together as much as possible.
  
  data representation for each square is: sb{index}({type}, ({x}, {y}), {replication}, {rotation}).
  
  For example if the square is sb1(N, (2, 3), 4, True) it means that this is the first group of squares in the grid, for which the type is N type, the starting square is located at [2,3] and the same square has been replicated 4 times. 
  since the rotation factor is true, that means the whole group is rotated by 90 degrees in the plane.
  
  Now, we have a sequential data of around 40 groups to place such that symmetry, and clustering and non overlapping criteria is preserved. 
  Each group which is placed after the first one is trying to maintain the above properties.
  
  we are giving you three examples, and want you to learn based on the data to give us the next group in the sequence:
  
  ****Example 1 ****: 
  
  ['sb1(P, (5, 17), 1, True)', 'sb2(P, (14, 17), 1, True)', ..., 'sb36(N, (11, 15), 1, True)']
  
  ****Example 2 ****: ...
  
  ****Example 3 ****: ... 
  
  Now, given these examples, here are the real datapoints that you need to learn:
  
  'sb1(P, (4, 19), 1, True)'"
  \end{tcolorbox}
  \caption{Sample training data for the finetuning for layout generation experiment: Starting from the first transistor, the model is finetuned to place transistors sequentially on a grid, 
  accounting for the constraints given in the prompt.}
  \label{fig:layout_finetune}
  \end{figure}

  \subsection{Fine-tuning Mistral7B with Synthetic Data v2 Experiment Details: }
  
  Fig~\ref{fig:promptv2} shows a sample of the full prompt that was used to finetune a Mistral-7B model:
  
  \begin{figure}[!htbp]
  \centering
  \begin{tcolorbox}[
    colback=gray!5, colframe=black!50,
    listing only,
    listing options={basicstyle=\ttfamily\tiny, breaklines=true},
    width=\columnwidth, 
  ]
  "We are performing a transistor placement task. The main goal here is to place transistors in a way that they do not overlap, are symmetrical with respect to y axis as much as possible, and grouped together as much as possible to minimize space. 
  Now, given the id which starts with sb, type which is either P or N , width, and height for a sequence of transistors, place them one by one such that these conditions are met., and give cx and cy in this format: (cx, xy). here are some examples:
  
  **** Example one ****
  
  Placed transistors: [] , Now place: [sb1, P, (?,?), 8, 1]$\rightarrow$ output: (5,11)
  
  Placed transistors: [sb1, P, (5,11), 8, 1] , Now place: [sb2, P, (?,?), 8, 1]$\rightarrow$ output: (15,11)
  
  Placed transistors: [sb1, P, (5,11), 8, 1, sb2, P, (15,11), 8, 1] , Now place: [sb3, N, (?,?), 6, 1]$\rightarrow$ output: (7,9)
  
  Placed transistors: [sb1, P, (5,11), 8, 1, sb2, P, (15,11), 8, 1, sb3, N, (7,9), 6, 1] , Now place: [sb4, N, (?,?), 6, 1]$\rightarrow$ output: (13,9)
  
  **** Example Two ****: ...
  
  **** Example Three ****: ...
  
  , Now lets place transistors,  Placed transistors: [] , Now place: [sb1, P, (?,?), 9, 1]$\rightarrow$ output: (4,4)"
  
  \end{tcolorbox}
  \caption{Sample training data for the finetuning for layout generation experiment v2: Starting from the first transistor, the model is finetuned to place transistors sequentially on a grid, 
  given the height and width and outputting the centerpoints accounting for the constraints given in the prompt.}
  \label{fig:promptv2}
  \end{figure}
  
  \subsection{GPT-oss-20B Model Outputs: }
  
  When prompting the fine-tuned GPT20B model, the prediction may contain detailed reasoning and step-by-step placement logic, as shown in Fig~\ref{fig:output_promptv3}:
  
  \begin{figure}[!htbp]
  \centering
  \begin{tcolorbox}[
    colback=gray!5, colframe=black!50,
    listing only,
    listing options={basicstyle=\ttfamily\tiny, breaklines=true},
    width=\textwidth, 
  ]
  We need to assign placements. We have 21 transistors? let's count: from list: P10,9,8,7,6 (5), N11-6 (7? actually N11,10,9,8,7,6,5 (7)) total12, N4,3,2,1,0 (5) total17, P11,5,4,3,2,1,0 
  (7) total24. Wait count again: after P types: have 5 Pshort, 7 Nshort, 5 nthick, 7 pthick = 24. Good.
  
  Goal place them maybe in two rows: P short near left, N short near left? maybe separate by type groups.
  We'll create coordinates $c_{x}$, $c_{y}$. Let's define grid spacing: horizontal spacing 0.5 (since widths up to 4). vertical stacking.
  
  Let's group as layers: top row: Pshort? Actually short have small widths 1.12 etc. Might place all pshort in one area. Nshort small too. 
  nthick bigger 1.11 etc. pthick largest up to 4.31. Let's use arrangement:
  Define origin at (0,0). We'll stack columns by type. We'll place Pshort leftmost, then Nshort, then nthick, then pthick.
  
  Let's compute bounding boxes per type. Pshort widths 1.12 height ~1.64/1.36. place in column with x starting at 0? We'll set $c_{x}$ = 0 + width/2? 
  but layout coordinates center? usually location of center? We'll set $c_{x}$ as center coordinate.
  Place them vertically stacked. Calculate Y positions: start y=0. For each transistor i, place y center at current y + height/2. 
  Then increment current y += height + vertical spacing (0.1?). 
  But we need avoid overlapping. Use 0.1 spacing. Let's create list. I'll prepare python to compute. But manual? I'll approximate. 
  First group Pshort: 5 items: P10 height 1.64 width1.12. center at y=1.64/2=0.82. Next P9 same height 1.64: next y center 0.82 + 1.64/2 +0.1+? 
  Actually need new center = previous y + previous height/2 + spacing + new height/2. 
  Let's just cumulative. Let's compute using simple: current y=0. for each transistor: $c_{y}$ = current y + h/2; current y += h + 0.1Ok. 
  I'll compute: trans list Pshort: 1. P10 h=1.64,$c_{y}$=0+0.82=0.82,current y=0+1.64+0.1=1.74 2. P9 h=1.64,$c_{y}$=1.74+0.82=2.56,current y=1.74+1.64+0.1=3.48 3. 
  P8 h=1.36,$c_{y}$=3.48+0.68=4.16,current y=3.48+1.36+0.1=4.94 4. P7 h=1.36,$c_{y}$=4.94+0.68=5.62,current y=4.94+1.36+0.1=6.40 5. 
  P6 h=1.64,$c_{y}$=6.40+0.82=7.22,current y=6.40+1.64+0.1=8.14 Okay. Same column $c_{x}$ = width/2 = 0.56? Actually width 1.12 $\rightarrow$ center 0.56. 
  use for all Pshort. Now Nshort group (7 trans): each width 0.44 height 0.7. current y start after previous group? maybe start from top of previous group? 
  We can place Nshort above Pshort column, but same x? Let's choose separate column with $c_{x}$ = 0.44/2=0.22? but they might overlap vertical if same column? But Pshort height sum 
  about 8.14. Nshort start at 0? Let's just start new column at x = 0.56 + 1.2 = 1.76 maybe. Let's use $c_{x}$ = 1.0? Wait we want separate columns. Let's set Pshort column x=0.56. 
  Nshort column x=0.56+1.0=1.56. Place Nshort with same algorithm: current y=0 1. N11 h=0.7 $\rightarrow$ $c_{y}$=0.35; current y=0+0.7+0.1=0.8 2. N10: $c_{y}$=0.8+0.35=1.15; 
  current y=0.8+0.7+0.1=1.6 3. N9: $c_{y}$=1.6+0.35=1.95; current y=1.6+0.7+0.1=2.4 4. N8: $c_{y}$=2.4+0.35=2.75; current y=2.4+0.7+0.1=3.2 5. 
  N7: $c_{y}$=3.2+0.35=3.55; current y=3.2+0.7+0.1=3. ~? 3.2+0.8=4.0 6. N6: $c_{y}$=4.0-? Wait after previous, current y=4.0 . $c_{y}$=4.0+0.35=4.35; current y=4.0+0.7+0.1=4.8 7. 
  N5: $c_{y}$=4.8+0.35=5.15; current y=4.8+0.7+0.1=5.6 Ok. Now nthick group width 1.11 height 1.72. For 5 trans (N4 to N0). We'll set column x maybe 1.56+1+? Suppose $c_{x}$=1.56+1.1=2.66? 
  Wait width 1.11 $\rightarrow$ center 0.555. Let's set column x=2.7 maybe. current y=0 1. N4 h 1.14 width 0.86? Wait N4 has width 0.86 height1.14 type nthick. Ok. So width different. 
  But they are all nthick; we can keep them in column. So for each transistor compute center accordingly. Set start at 0. 1. N4 w0.86 h1.14 $c_{x}$=0.43. 
  $c_{y}$=0+1.14/2=0.57. current y=0+1.14+0.1=1.24 2. N3 w1.11 h1.72 $c_{y}$=1.24+0.86=2.10? Wait 1.72/2=0.86. So $c_{y}$=1.24+0.86=2.10. current y=1.24+1.72+0.1=2. 
  (1.24+1.82=3.06?) Let's compute: 1.24+1.72=2.96; +0.1=3.06 3. N2: $c_{y}$=3.06+0.86=3.92; current y=3.06+1.72+0.1=4.88 4. N1: $c_{y}$=4.88+0.86=5.74; current y=4.88+1.72+0.1=6.70 5. 
  N0: $c_{y}$=6.70+0.86=7.56; current y=6.70+1.72+0.1=8.52 All with $c_{x}$ for each: N4 0.43; others 0.555. Now pthick group: widths up to 4.31 height 4.36. 7 trans. 
  We'll put them in column maybe x=4.0? Let's compute. Set $c_{x}$ for each = width/2. Start current y=0. 1. P11 w4.31 h4.36 $c_{x}$=2.155. $c_{y}$=4.36/2=2.18. 
  current y=0+4.36+0.1=4.46 2. P5 w2.48 h3.2 $c_{x}$=1.24. $c_{y}$=4.46+3.2/2=4.46+1.6=6.06. current y=4.46+3.2+0.1=7.66 3. P4 w2.75 h4.36 $c_{x}$=1.375. $c_{y}$=7.66+2.18=9.84. 
  current y=7.66+4.36+0.1=12.12 4. P3 w2.75 h4.36 $c_{x}$=1.375. $c_{y}$=12.12+2.18=14.30. current y=12.12+4.36+0.1=16.58 5. P2 w2.48 h10.36 $c_{x}$=1",
  \end{tcolorbox}
  \caption{Sample full output after prompting the GPT-oss-20B. Obviously, parsing the center-points is non-trivial.}
  \label{fig:output_promptv3}
  \end{figure}

  \clearpage
  \bibliography{references.bib}

\end{document}